\documentclass[conference]{IEEEtran}
\IEEEoverridecommandlockouts
\usepackage[noadjust]{cite}
\usepackage{amsmath,amssymb,amsfonts}
\usepackage{algorithmic}
\usepackage{graphicx}
\usepackage{textcomp}
\usepackage{xcolor}
\usepackage[utf8]{inputenc}
\usepackage[vietnamese,main=english]{babel}
\def\BibTeX{{\rm B\kern-.05em{\sc i\kern-.025em b}\kern-.08em
    T\kern-.1667em\lower.7ex\hbox{E}\kern-.125emX}}
    
\newcommand{\VI}{\foreignlanguage{vietnamese}}
    
\usepackage{booktabs} 
\usepackage{multirow}   
    
\begin{document}

\title{Attentive Neural Network for Named Entity Recognition in Vietnamese
}

\author{
\IEEEauthorblockN{\textsuperscript{1,2}Kim Anh Nguyen}
\IEEEauthorblockA{\textsuperscript{1}\textit{FPT Technology Research Institute} \\
\textsuperscript{2}\textit{Hung Vuong University}\\
anhnk14@fpt.com.vn}
\and
\IEEEauthorblockN{\textsuperscript{1}Ngan Dong}
\IEEEauthorblockA{\textsuperscript{1}\textit{FPT Technology Research Institute} \\
ngandt3@fpt.com.vn}
\and
\IEEEauthorblockN{\textsuperscript{3}Cam-Tu Nguyen}
\IEEEauthorblockA{\textsuperscript{3}\textit{National Key Laboratory for} \\
\textit{Novel Software Technology} \\
\textit{Nanjing University}\\
ncamtu@nju.edu.cn}
}

\maketitle

\begin{abstract}
We propose an attentive neural network for the task of named entity recognition in Vietnamese. The proposed attentive neural model makes use of character-based language models and word embeddings to encode words as vector representations. A neural network architecture of encoder, attention, and decoder layers is then utilized to encode knowledge of input sentences and to label entity tags. The experimental results show that the proposed attentive neural network achieves the state-of-the-art results on the benchmark named entity recognition datasets in Vietnamese in comparison to both hand-crafted features based models and neural models.
\end{abstract}


\section{Introduction}
\label{sec:intro}
Named entity recognition (NER) is one of fundamental sequence labeling tasks as well as other tasks such as word segmentation, part-of-speech (POS) tagging, or noun phrase chunking. The NER task aims to identify named entities in the given texts and then to assign named entities to particular entity types such as location, organization or person name. NER task plays a crucial role in natural language understanding and downstream applications such as relation extraction, entity linking, question answering, or machine translation. 

In the previous studies, NER approaches make use of linear statistical models to label entity tags such as hidden Markov models (HMM), maximum entropy models (ME), or conditional random fields (CRF) (\cite{Lafferty:EtAl:2001}). However, most those kinds of models rely heavily on hand-crafted features and task-specific resources, leading that those models are difficult to adapt to new tasks or to shift to new domains. For example, in English, orthographic features and external resources of gazetteers are commonly used in NER task. For Vietnamese, the approach in \cite{Minh:2018, Pham:vlsp:2018} used the information of word, word shapes, part-of-speech tags, chunking tags as hand-crafted features for CRF to label entity tags.

In the past few years, neural networks for NER have been proposed to deal with drawbacks of statistical-based NER models by extracting automatically features instead creating heavily hand-crafted features. Neural architectures for NER often make use of the combination of either recurrent neural network (RNN) and CRF or convolution neural network (CNN) and CRF to extract automatically information from the inputs and detect NER labels. Reference \cite{Lample:EtAl:2016}, among others, proposed a neural architecture by using recurrent neural network with long short-term memory units (LSTM) (\cite{Hochreiter/Schmidhuber:1997}) and CRF to label NER tags. Moreover, the combination of bidirectional LSTM, CNN, and CRF is introduced to obtain benefits from both word- and character-level representations automatically for detecting NER labels as in \cite{Ma/Hovy:2016}. Recently, as in \cite{Liu:EtAl:2018}, a combination of language model (LM), LSTM, and CRF is used to extract knowledge from raw texts and empower the sequence labeling task including NER task. For Vietnamese, a non-hand-crafted feature based model which is combination of LSTM, CNN, and CRF is applied to solve the task of Vietnamese NER as in \cite{Pham:EtAl:2017}. Moreover, ZA-NER model (\cite{Luong/Pham:2018}) which is based on a combination of bidirectional LSTM and CRF is proposed to extract named entities. 

In this paper, we introduce an attentive neural network (VNER) for Vietnamese NER task without using any hand-crafted features or task-specific resources. In the proposed neural network, we incorporate a neural language model to encode the character-based words. Similar to \cite{Liu:EtAl:2018}, the prediction of the next character in the language model is adapted to predict the next word. Moreover, the pre-trained word embeddings are also utilized to extract knowledge from word level. The concatenation of character-based word and pre-trained word embedding is then used as the vector representation of a word- or token-layer. A bidirectional LSTM is then applied as an encoder layer to encode the knowledge of the input sentence. We then make use of a LSTM as a decoder together with an attention mechanism to decode the outputs of encoder layer. Finally, a CRF layer is used to model context dependencies and entity labels.

For the experiment, we evaluate the VNER model on two benchmark datasets of Vietnamese NER task which are VLSP-2016 (\cite{Huyen/Luong:2016}) and VLSP-2018\footnote{http://vlsp.org.vn/vlsp2018/eval/ner} NER datasets. The experimental results show that the VNER model achieves the state-of-the-art results compared to both hand-crafted based models and neural models.

\section{Related Work}
\label{sec:related-work}
\begin{figure*}[t]
    \centering
    \resizebox{.7\linewidth}{!}{
        \includegraphics{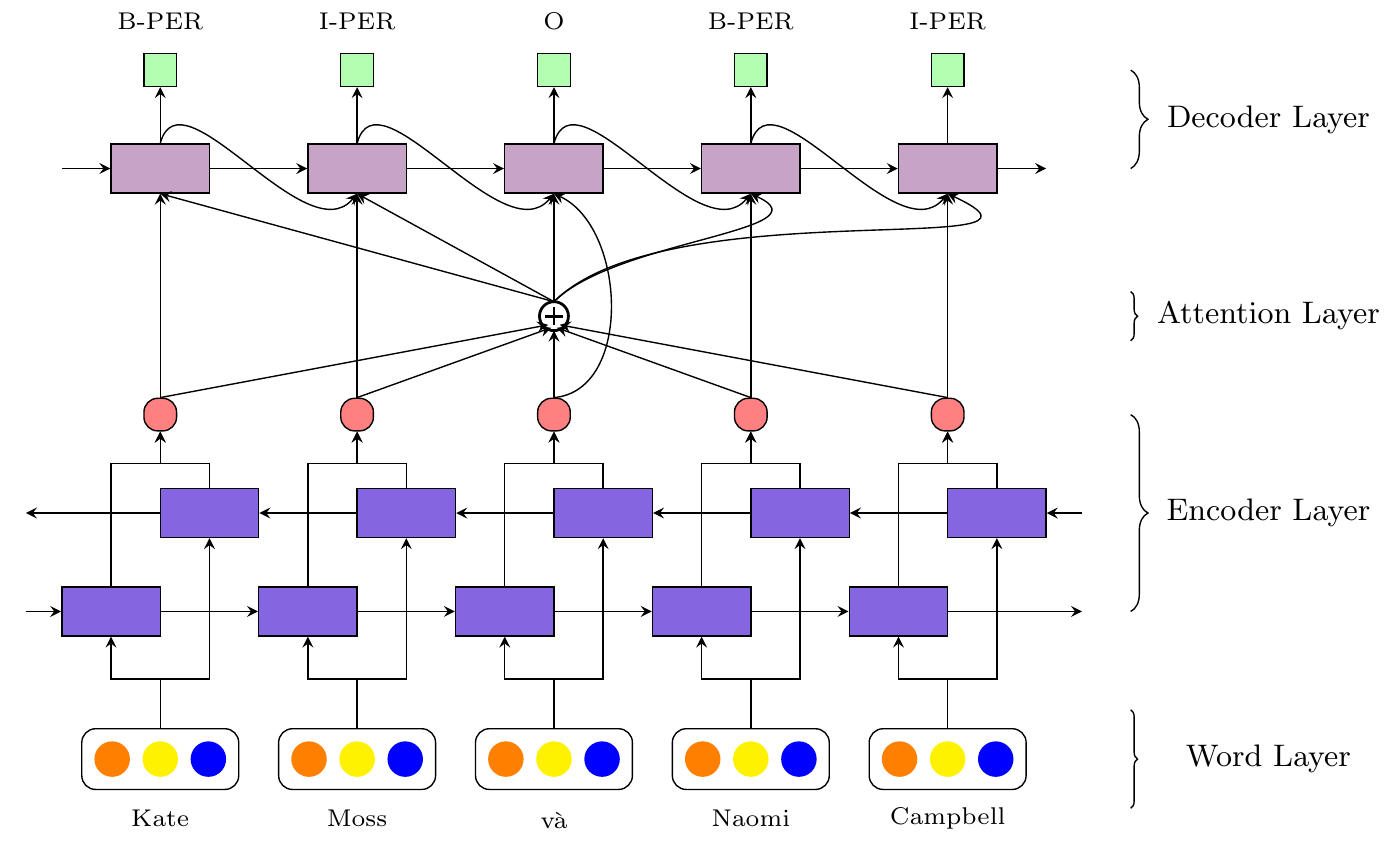}
    }
    \caption{The architecture of VNER model}
    \label{fig:architecture}
\end{figure*}
Named entity recognition is a fundamental NLP research problem that has been studied for years. In the literature, proposed approaches for NER task can be divided into two types: the first type is based on linear statistical approaches and the second type is based on neural models.

In the first approach, based-statistical NER systems have been dominated for years. These systems rely on a pre-defined set of hand-crafted features such as lemma, word embeddings, semantic dictionary, word-shape, POS tags, or chunking tags. Each sentence is represented as a set of features and then fed into a linear model such as HMM, ME, or CRF to label entity tags for each word or token. In comparison with neural models, this type of NER systems is straightforward and requires less resources. In addition, these system are also proved to work well for low-resource languages such as Vietnamese. However, these kinds of NER systems are relied heavily on the feature set use, and on hand-crafted features that are expensive to construct and are difficultly reusable.

For the second approach, thanks to the recent advancements in computing technology, neural-based models have emerged as a powerful tools for a number of research problems, including sequence labeling tasks. Neural-based NER systems are end-to-end systems that require no exclusively pre-defined features. Those proposed models are based on complex deep learning architecture such as RNN, LSTM, or CNN (\cite{Lample:EtAl:2016, Ma/Hovy:2016,Liu:EtAl:2018,Pham:EtAl:2017}). Word embeddings and/or character embeddings are often used to represent the semantic relations of words or characters. Other information such as POS tags or chunking tags are also used to provide additional syntactic information. Sentences are represented as vector representations and fed into  variety architectures of deep neural networks to encode knowledge from them. A CRF layer then can be used on top to infer entity tags for words or tokens. Consequently, neural models are easy to adapt to new domains and can achieve state-of-the-art results on many sequence labeling problems. However, due to this type of models that is quite complex, these models require large training data and take time for training.

There have been considerable work proposed by Vietnamese researchers in solving the NER problem such as dynamic feature induction model (\cite{Vu:EtAl:2018}), CRF model (\cite{Minh:2018}), or LSTM (\cite{Luong/Pham:2018,Pham:EtAl:2017}). CRF-based model achieves state-of-the-art results on the VLSP 2016 and VLSP 2018 competitions; however, it still suffers from the linear statistical model drawbacks as mentioned above. Our proposed attentive neural network has similar architecture as the ones mentioned in \cite{Luong/Pham:2018,Pham:EtAl:2017} with additional highway layers for enhancing word embeddings and character embeddings at run time. Moreover, an attention mechanism is used to further improve the system performance. 

\section{VNER: An Attentive Neural Network for Vietnamese Named Entity Recognition}
\label{sec:vner}
In this section, we describe the components (layers) in the architecture of VNER model. The neural architecture of the proposed model is visualized in Figure \ref{fig:architecture}. The VNER model includes the layers of word, encoder, attention, and decoder.

\subsection{Word Layer}
\label{subsec:word-layer}
\begin{figure*}[h]
    \centering
    \resizebox{.7\linewidth}{!}{
    \includegraphics{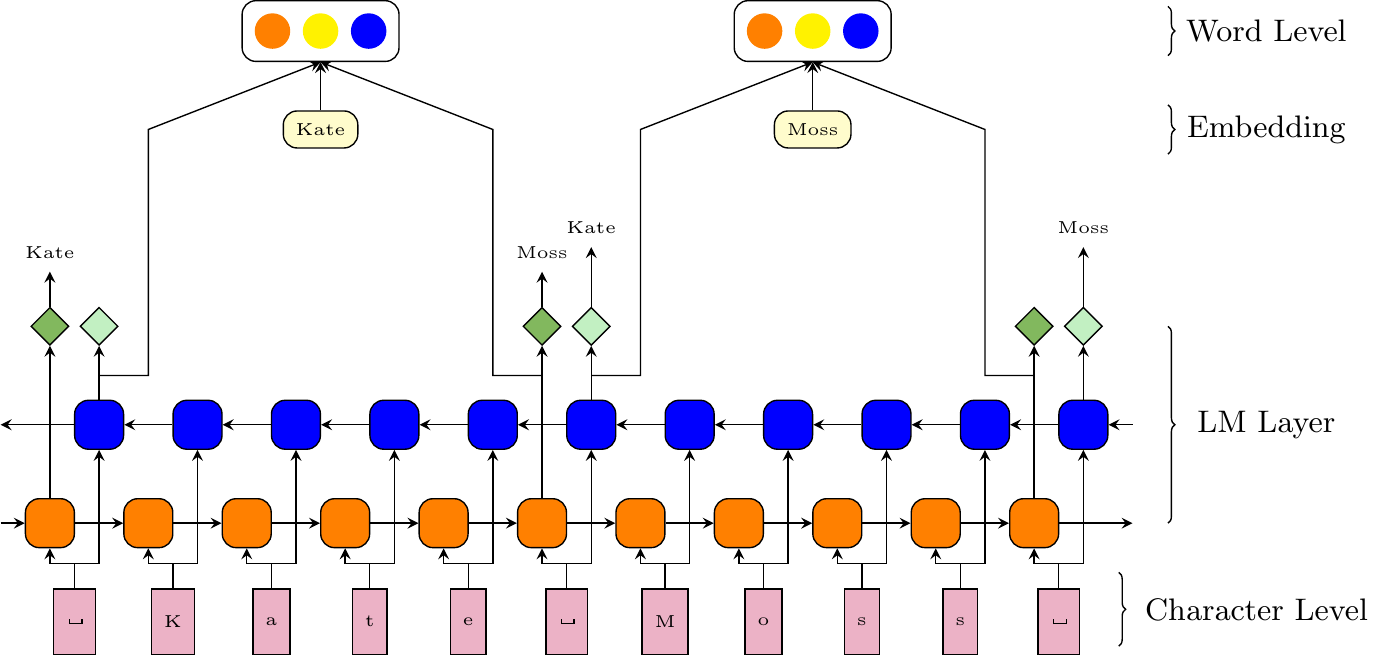}
    }
    \caption{The illustration for constructing word layer}
    \label{fig:word_layer}
\end{figure*}
As mentioned in Section \ref{sec:intro}, a word is the concatenation of both character- and word-based vector representations. The visualization of word layer is illustrated in Figure \ref{fig:word_layer}. Specifically, for the character-based word, we adopt character-level layer in \cite{Liu:EtAl:2018} to represent the vector representation of character-based word by using the neural language model. The character-level language models are trained on unannotated sequence of the input sentence. Furthermore, we make use of two LSTM networks to model the sequence of characters in both forward and backward directions. For each word in the input sentence, the prediction of all characters in each word is the next word, helping to capture better lexical information of the next word rather than its spelling. 

In addition, we employ four highway units (\cite{Srivastava:EtAl:2015}) to enhance the output's performance of language model. Specifically, the highway units compute non-linear transformation as follows:
\begin{equation}
	\mathbf{O} = H(\mathbf{x}) = t \odot \Phi(\mathbf{W}_{H} \mathbf{x} + b_H)
\end{equation}
where $\Phi$ is non-linear activation function, $\odot$ is the element-wise product, and $t = \sigma (\mathbf{W}_{T} \mathbf{x} + b_T)$ is transform gate. In our model, four highway units are applied to both forward and backward directions of LSTM networks. The two first highway units transfer the output of forward LSTM network $\mathbf{O}_{f_i}$ to $\mathbf{O}^{L}_{f_i}$ and $\mathbf{O}^{N}_{f_i}$, in which $\mathbf{O}^{L}_{f_i}$ and $\mathbf{O}^{N}_{f_i}$ are used in forward language model and word layer, respectively. Similarly, the two transformations of the two last highway units that transfer $\mathbf{O}_{b_i}$ to $\mathbf{O}^{L}_{b_i}$ and $\mathbf{O}^{N}_{b_i}$ are used in backward language model and word layer, respectively.

Finally, the vector representation of a word $w_i$ is the concatenation of its pre-trained embedding $\mathbf{E}_i$, the forward transformation of highway units $\mathbf{O}^{N}_{f_i}$, and the backward transformation of highway units $\mathbf{O}^{N}_{b_{i-1}}$.
\begin{equation}
	\vec{w}_i = [\vec{\mathbf{E}}_i; \vec{\mathbf{O}}^{N}_{f_i}; \vec{\mathbf{O}}^{N}_{b_{i-1}}]
\end{equation}

\subsection{Encoder Layer}
\label{subsec:encoder-layer}
In order to capture the information of whole input sequence, we make use of bidirectional LSTM networks to encode an input sequence. Precisely, for a given input sentence $\mathbf{x} = (\mathbf{x_1, x_2,..., x_n})$ containing $n$ words. A forward LSTM network aims to encode the input sequence $\mathbf{x}$ from the start to the end of the input sequence and to generate a hidden representation $\vec{\mathbf{h}}^f_t$ at every word $t$; and a backward LSTM network encodes the input sequence from the end to start of the input sequence and computes a hidden representation $\vec{\mathbf{h}}^b_t$ at every word $t$.

The representation of a word $\mathbf{w}_t$ in the input sentence $\mathbf{x}$ using the bidirectional LSTM encoder is then obtained by concatenating its forward hidden representation and backward hidden representation as follows:
\begin{equation}
	\vec{\mathbf{h}}_t = [\vec{\mathbf{h}}^f_t; \vec{\mathbf{h}}^b_t]
\end{equation}

\subsection{Attention Layer}
\label{subsec:attention-layer}
The attention mechanism has gained popularity in recent years in training neural networks. Reference \cite{Bahdanau:EtAl:2015} proposed and successfully applied attention mechanism to jointly translate and align words for the task of neural machine translation. Ideally, attention mechanism is often applied in between the layers of encoder and decoder, aiming to selectively focus on parts of encoder layer's outputs corresponding to time step in the decoder layer. Specifically, the attention layer first takes the outputs of encoder layer as inputs to compute probability distribution of encoder's outputs for each word $w_t$ at the time step $t$ of decoder layer as follows:
\begin{align}
	score(h_t, \bar{h}_e) &= h_t \cdot \bar{h}^{\mathtt{T}}_e \\
	\alpha_t(e) &= \frac{ \exp (score(h_t, \bar{h}_e))}{\sum \nolimits_{e'} \exp (score(h_t, \bar{h}_e'))} \\
	c_t &= \sum \limits_{e} \alpha_t(e) \bar{h}_e
\end{align}
in which $h_t$ is hidden state of decoder layer at time step $t$; $\alpha_t(e)$ is the attention weights according to the hidden states of encoder layer $\bar{h}_e$ and $h_t$; and $c_t$ is the context vector of attention layer.

\subsection{Decoder Layer}
\label{subsec:decoder-layer}
This layer aims to decode and label entity tags which are dependency tags. Therefore, it is beneficial to observe the relationships between entity tags in neighborhoods and jointly decode the highest probability of entity tags for a given input sentence. For example, it is meaningless to label \texttt{I-PER} after \texttt{I-ORG} in the NER task with BIO annotation.
To do so, firstly, we make use of a LSTM network which is considered as a decoder to process the outputs of encoder and attention layers. Given the outputs of encoder layer $h_e = [h_{e_1}, h_{e_2},..., h_{e_n}]$, the input of decoder layer at each time step $t$ is the concatenation of three elements which are the hidden state $h_{e_t}$ of encoder layer, the previous state of decoder layer $h_{d_{t-1}}$, and the context vector of attention layer $c_t$ as follows:
\begin{equation}
	h_{d_t} = [h_{e_t}; h_{d_{t-1}}; c_t]
\end{equation}

Secondly, we apply CRF model to the outputs of the decoder to model dependency tags. Formally, as in \cite{Ma/Hovy:2016}, we use $\mathbf{z = (z_1, z_2,..., z_n})$ to represent the output of the decoder in which $\mathbf{z_i}$ is the vector representation of $i$th word in the input sentence; $\mathcal{Y}(\mathbf{z})$ stands for the set of all possible sequences for $\mathbf{z}$; and $\mathbf{y = (y_1, y_2,..., y_n})$ represents the sequence of labels for $\mathbf{z}$. The probabilities of possible label sequences $\mathbf{y}$ given $\mathbf{z}$ are defined as follows:
\begin{equation}
	p(\mathbf{y}|\mathbf{z}; \mathbf{W,b}) = 
	\frac{\prod \limits_{i=1}^{n} {\psi_i(y_{i-1},y_i,\mathbf{z})}}{\sum \limits_{y' \in \mathcal{Y(\mathbf{z})}} \prod \limits_{i=1}^{n} {\psi_i(y'_{i-1},y'_i,\mathbf{z})}} \label{eq:crf}
\end{equation}
where $\psi_i(y',y,\mathbf{z}) = \exp (\mathbf{W}^{\mathtt{T}}_{y',y} \mathbf{z}_i + \mathbf{b}_{y',y})$ are potential functions; and $\mathbf{W}^{\mathtt{T}}_{y',y}$ and $\mathbf{b}_{y',y}$ are weight matrix and bias corresponding to the label pair $(y',y)$, respectively.

\subsection{Joint Training}
\label{subsec:joint-training}
As mentioned in Section \ref{subsec:word-layer}, in the VNER model,  we make use of both forward and backward neural language models to jointly learn character-based word embeddings. Specifically, due to both neural language models which consider the predictions to words and utilize the character sequence as the inputs, the probabilities for the models of both forward $p_f$ and backward $p_b$ to generate words are defined as follows:
\begin{align}
	p_f(x_1,..., x_n) = \prod \limits_{i=1}^{n} p_f (x_i | c_{0,\_},..., c_{i-1,\_}) \label{eq:flm} \\ 
	p_b(x_1,..., x_n) = \prod \limits_{i=1}^{n} p_b (x_i | c_{i+1,\_},..., c_{n,\_}) \label{eq:blm}
\end{align}
where $x_i$ is the $i$th word; and $c_{i,j}$ is the $j$th character of the $i$th word.

By combining the equations of \ref{eq:crf}, \ref{eq:flm}, and \ref{eq:blm}, the objective function of joint model used to label NER tags can be defined as the following equation:
\begin{equation}
	\mathcal{J} = - \sum \limits_{i} \Big ( p (\mathbf{y}_i|\mathbf{z}_i ) + \lambda ( \log p_f (\mathbf{x}_i) + \log p_b (\mathbf{x}_i) ) \Big )
\end{equation}
in which $\lambda$ is a hyper-parameter.

\section{Experiments}
\label{sec:experiments}

\begin{table}[h!]
\caption{The size of VLSP-2016 and VLSP-2018 datasets}
\label{tbl:dataset-size}
\centering
\begin{tabular}{|l|rr|rrr|}
\hline
\multirow{2}{*}{\textbf{Type}} & \multicolumn{2}{c|}{\textbf{VLSP-2016}}                                 & \multicolumn{3}{c|}{\textbf{VLSP-2018}}                                                                    \\ \cline{2-6} 
                               & \multicolumn{1}{l}{\textbf{Train}} & \multicolumn{1}{l|}{\textbf{Test}} & \multicolumn{1}{l}{\textbf{Train}} & \multicolumn{1}{l}{\textbf{Test}} & \multicolumn{1}{l|}{\textbf{Dev}} \\ \hline
LOC                            & 6,245                              & 1,379                              & 8,831                              & 2,525                             & 3,043                             \\
ORG                            & 1,213                              & 274                                & 3,471                              & 1,616                             & 1,203                             \\
PER                            & 7,480                              & 1,294                              & 6,427                              & 3,518                             & 2,168                             \\
MISC                           & 282                                & 49                                 & 805                                & 296                               & 179                               \\
 \hline
\end{tabular}
\end{table}

\subsection{NER datasets}
\label{subsec:datasets}
In our experiments, we evaluate the VNER model on two benchmark datasets for Vietnamese NER which are VLSP-2016 NER task and VLSP-2018 NER task. Specifically, VLSP-2016 dataset consists of four entity types including location (LOC), organization (ORG), person (PER), and miscellaneous (MISC). VLSP-2016 dataset additionally provides the information about word segmentation, part-of-speech, and chunking tags. Similarly, VLSP-2018 dataset also contains four types of entity which are LOC, ORG, PER, and MISC. However, unlike VLSP-2016 dataset, VLSP-2018 dataset is only annotated without having any additional information such as word segmentation, POS, or chunking tags. In addition, VLSP-2018 dataset contains nested entities which contain other entities inside them. In this paper, we conduct experiments on single-layer for VLSP-2016 dataset. In that case, we retain only the entities tags of the outer-most level. For the VLSP 2018 dataset, we experiment our model on both single-layer and nested entities. Table \ref{tbl:dataset-size} shows the statistic of our two experimental NER datasets.

\subsection{Experimental Settings}
\label{subsec:settings}
Due to VLSP-2016 dataset that does not have development set, hence we create a development set by sampling randomly 2000 samples of train set as in \cite{Vu:EtAl:2018}; and the rest of train set is used for training VNER model.
We then train VNER model on both VLSP-2016 and VLSP-2018 datasets with the train set and further tune up the model with the development set. The parameters used to train VNER model are summarized in Table \ref{tbl:model-params}.

\begin{table}[h!]
\caption{The model parameters}
\label{tbl:model-params}
\centering
\begin{tabular}{|l|r|}
\hline
\textbf{Param} & \textbf{Value} \\ \hline                                
character hidden dim. & 300 \\ 
word hidden dim. & 300
\\
char dim. & 30 
\\
word dim. & 100 
\\
dropout rate & 0.6 
\\
No. of word layers & 1 
\\
No. of char layers & 1 
\\
update function & Adam 
\\
learning rate & 0.001 
\\
batch size & 128 
\\
\hline
\end{tabular}
\end{table}

\subsection{Results}
\label{subsec:results}
\begin{table*}[t]
\caption{The performance of VNER model compared to baseline models on VLSP-2016 dataset}
\label{tbl:vlsp2016:results}
\centering
\begin{tabular}{|l|ccccc|}
\hline
\multicolumn{1}{|c|}{\textbf{Setting}} & \textbf{VNER}  & \textbf{Feature-based CRF} & \textbf{VnCoreNLP} & \textbf{NNVLP} & \textbf{vie-ner-lstm} \\ \hline
Without POS, chunking tags (1st setup)            & \textbf{90.37} & 90.03                      & -                  & -              & -                \\
Annotated POS, chunking tags (2nd setup)         & \textbf{95.33} & 93.93                      & -                  & 92.91          & 92.05            \\
Underthesea-based POS, chunking tags (3rd setup)   & \textbf{90.17} & 89.30                      & -                  & -              & -                \\
VnCoreNLP setup                        & \textbf{89.58} & -                          & 88.55              & -              & -                \\ \hline
\end{tabular}
\end{table*}

For VLSP-2016 dataset, we compare VNER model with CRF model based on hand-crafted features (\cite{Minh:2018, Pham:vlsp:2018}) (henceforth, feature-based CRF) which is the state-of-the-art model on this dataset. To do so, we evaluate VNER model on three setups as reported in \cite{Minh:2018}. In the first setup, we use only word information to train VNER model without using the information of annotated POS and chunking tags. For the second setup, we train VNER model by using all annotated information including word, POS tags, and chunking tags. For the last setup, we rely on the Underthesea toolkit to generate POS and chunking tags for the sentences in VLSP-2016 dataset. We then make use of the information of word, those generated POS, and chunking tags to train VNER model. Moreover, we also evaluate VNER model on VLSP-2016 dataset with another setup as in the experiment of VnCoreNLP (\cite{Vu:EtAl:2018}) in which the contiguous syllable constituting a PER tag is merged to form a word. In comparison to other neural network for Vietnamese NER task, we compare the performance of VNER model with two neural models: NNVLP model (\cite{Pham:EtAl:2017}) that makes use of the combination of bidirectional LSTM, CNN, and CRF models; and vie-ner-lstm model (\cite{Pham:EtAl:2017b}) that incorporates automatic syntactic features with word embeddings as input for bidirectional LSTM network.

\begin{table}[!]
\caption{Examples of joint-tags for 2 levels}
\label{tbl:joint:tags}
\centering
\begin{tabular}{|l|ccc|}
\hline
\multicolumn{1}{|c|}{\textbf{Token}} & \textbf{Level-1 Tag} & \textbf{Level-2 Tag} & \textbf{Joint Tag}    \\ \hline
Ca                              & O              & O           & O$+$O
\\
\VI{mổ}                             & O              & O           & O$+$O
\\
do                              & O              & O           & O$+$O
\\
\VI{bác}                              & O              & O           & O$+$O
\\
\VI{sĩ}                             & O              & O           & O$+$O
\\
T.N.Q.P                              & O              & B-PER           & O$+$B-PER
\\
\VI{thực}                              & O              & O           & O$+$O
\\
\VI{hiện}                            & O              & O           & O$+$O
\\
.                              & O              & O           & O$+$O
\\\hline
\end{tabular}
\end{table}

Table \ref{tbl:vlsp2016:results} shows the performance of VNER model and baseline models in terms of F1 score. Note that all results of baseline models are reported from their original experiments. Overall, VNER model outperforms all baseline models across four setups of training data. In the first setup, VNER model obtains a comparable result compared to feature-based CRF model with 90.37 F1 score for VNER model in comparison to 90.03 F1 score for feature-based CRF model. For the second setup, four models used all information about word and the annotation of POS, and chunking tags to train those models. VNER model shows a large improvement with 95.33 F1 score compared to three baseline models including feature-based CRF, VnCoreNLP, and vie-ner-lstm with the F1 scores of 93.93, 92.91, and 92.05, respectively. Similar to three above setups, in the VnCoreNLP setup, VNER model also outperforms VnCoreNLP model where PER entity was reconstructed by merging contiguous syllables to a word form. Concretely, VNER model achieves 89.58 F1 score in comparison with 88.55 F1 score of VnCoreNLP. The motivation behind is to make the dataset to be more realistic. Because the annotated POS tags are not available in the real-world application.

For VLSP-2018 dataset, we conduct two experiments on both nested and single-layer entities. In the first experiment for single layer, we experiment the performance of VNER model in comparison with the performance of two baseline models which are feature-based CRF and ZA-NER models. While feature-based CRF model relied on variety of hand-crafted features, both VNER and ZA-NER models are trained on the original data of VLSP-2018 NER dataset. Furthermore, in order to observe the effect of attention layer on VNER model, we disable this layer and apply CRF model instead. Table \ref{tbl:vlsp2018:results} shows the performance of three models on recognizing named entities of VLSP-2018 NER dataset. Concretely, while VNER model outperforms both feature-based CRF and ZA-NER models with 77.52 F1 score of VNER model compared to 76.63 F1 score and 74.00 F1 score for feature-based CRF model and ZA-NER model, respectively, the VNER model without attention layer shows the low performance with 73.23 F1 score.

\begin{table}[!]
\caption{Performance of VNER and baseline models on VLSP-2018 dataset}
\label{tbl:vlsp2018:results}
\centering
\begin{tabular}{|l|ccc|}
\hline
\multicolumn{1}{|c|}{\textbf{Model}} & \textbf{Precision} & \textbf{Recall} & \textbf{F1}    \\ \hline
ZA-NER                               & 76.00              & 72.00           & 74.00
\\
Feature-based CRF                    & 73.46              & 80.08           & 76.63          \\
VNER w/o attention                                 & 75.19              & 71.37           & 73.23 \\
VNER                                 & 75.70              & 79.43           & \textbf{77.52} \\ \hline
\end{tabular}
\end{table}

In the second experiment for nested entities, we use the same experimental setup as run-\#4 in \cite{Minh:2018}. For each token, we define a joint tag which indicates a combination of both tags from level 1 and level 2. Table \ref{tbl:joint:tags} illustrates an example of joint tags used in our experiment. By doing so, our model is able to work with nested named entities. Table \ref{tbl:vlsp2018:joint:results} shows the performance of our system compared to Feature-based CRF model. In fact, VNER model outperforms Feature-based CRF by more than 3\% of F1 score.

\begin{table}[h]
\caption{Performance of two models for nested entities}
\label{tbl:vlsp2018:joint:results}
\centering
\begin{tabular}{|l|ccc|}
\hline
\multicolumn{1}{|c|}{\textbf{Model}} & \textbf{Precision} & \textbf{Recall} & \textbf{F1}    \\ \hline
Feature-based CRF                    & 77.99              & 77.1        & 74.7         \\
VNER                                 & 79.23              & 76.57           & \textbf{77.88} \\ \hline
\end{tabular}
\end{table}

\section{Conclusion}
\label{sec:conclusion}
This paper proposed an attentive neural model, namely VNER to recognize named entities in Vietnamese. The VNER model is constructed by four main layers including word layer, encoder layer, attention layer, and decoder layer. A series of experiments on benchmark datasets for Vietnamese NER task showed that VNER model outperform both hand-crafted feature and neural network models. 



\begin{thebibliography}{00}

\bibitem{Lafferty:EtAl:2001} John Lafferty, Andrew McCallum, and Fernando Pereira, ``Conditional Random Fields: Probabilistic Models for Segmenting and Labeling Sequence Data'', In Proceedings of the Eighteenth International Conference on Machine Learning (ICML), pp. 282--289, 2001.

\bibitem{Minh:2018} Pham Quang Nhat Minh, ``A Feature-Rich Vietnamese Named-Entity Recognition Model'', arXiv:1803.04375.

\bibitem{Pham:vlsp:2018} Pham Quang Nhat Minh, ``A Feature-Based Model for Nested Named-Entity Recognition at VLSP-2018 NER Evaluation Campaign'', In Proceedings of Vietnamese Speech and Language Processing (VLSP), 2018.

\bibitem{Lample:EtAl:2016} Guillaume Lample, Miguel Ballesteros, Sandeep Subramanian, Kazuya Kawakami, and Chris Dyer, ``Neural Architectures for Named Entity Recognition'', In Proceedings of The 2016 Conference of the North American Chapter of the Association for Computational Linguistics: Human Language Technologies (NAACL-HTL), pp. 260--270, 2016.

\bibitem{Hochreiter/Schmidhuber:1997} Sepp Hochreiter and J\"{u}rgen Schmidhuber, ``Long Short-Term Memory'', Neural Computation, vol. 9, pp. 1735--1780, 1997.

\bibitem{Ma/Hovy:2016} Xuezhe Ma and Eduard Hovy, ``End-to-end Sequence Labeling via Bi-directional LSTM-CNNs-CRF'', In Proceedings of the 54th Annual Meeting of the Association for Computational Linguistics (ACL), pp. 1064--1074, 2016.

\bibitem{Liu:EtAl:2018} Liyuan Liu, Jingbo Shang, Xiang Ren, Frank Fangzheng Xu, Huan Gui, Jian Peng, and Jiawei Han, ``Empower Sequence Labeling with Task-Aware Neural Language Model'', AAAI Conference on Artificial Intelligence, 2018. 

\bibitem{Pham:EtAl:2017} Thai-Hoang Pham, Xuan-Khoai Pham, Tuan-Anh Nguyen, and Phuong Le-Hong, ``NNVLP: A Neural Network-Based Vietnamese Language Processing Toolkit'', In The Companion Volume of the IJCNLP 2017 Proceedings: System Demonstrations, pp. 37--40, 2017.

\bibitem{Luong/Pham:2018} Luong Viet-Thang and Long Kim Pham, ``ZA-NER: Vietnamese Named Entity Recognition at VLSP 2018 Evaluation Campaign'', In Proceedings of Vietnamese Speech and Language Processing (VLSP), 2018.

\bibitem{Huyen/Luong:2016} Nguyen Thi Minh Huyen and Vu Xuan Luong, ``Vlsp 2016 shared task: Named entity recognition'', In Proceedings of Vietnamese Speech and Language Processing (VLSP), 2016.

\bibitem{Vu:EtAl:2018} Thanh Vu, Dat Quoc Nguyen, Dai Quoc Nguyen, Mark Dras, and Mark Johnson, ``VnCoreNLP: A Vietnamese Natural Language Processing Toolkit'', In Proceedings of NAACL-HLT 2018: Demonstrations, pp. 56--60, 2018.

\bibitem{Srivastava:EtAl:2015} Rupesh Kumar Srivastava, Klaus Greff, and J{\"{u}}rgen Schmidhuber, ``Highway Networks'', arXiv:1505.00387, 2015. 

\bibitem{Bahdanau:EtAl:2015} Dzmitry Bahdanau, Kyunghyun Cho, and Yoshua Bengio, ``Neural machine translation by jointly learning to align and translate'', arXiv:1409.0473, 2015.

\bibitem{Pham:EtAl:2017b} Thai-Hoang Pham and Phuong Le-Hong, ``The Importance of Automatic Syntactic Features in Vietnamese Named Entity Recognition'', In Proceedings of 31st Pacific Asia Conference on Language, Information and Computation (PACLIC 31), pp. 97--103, 2017.


\end{thebibliography}
\end{document}